\pdfoutput=1

\documentclass[11pt]{article}

\usepackage[]{acl}

\usepackage{times}
\usepackage{latexsym}

\usepackage[T1]{fontenc}

\usepackage[utf8]{inputenc}

\usepackage{microtype}

\usepackage{inconsolata}
\usepackage{graphicx}
\usepackage{url}
\usepackage{float} 
\usepackage{caption}
\usepackage{subcaption}

\usepackage{array,makecell}

\title{Improved Text Emotion Prediction Using Combined Valence and Arousal Ordinal Classification}

 \author{Michael Mitsios$^{\star}$, Georgios Vamvoukakis$^{\star}$, Georgia Maniati$^{\star}$, Nikolaos Ellinas$^{\star}$, Georgios Dimitriou$^{\star}$ \\
	{\bf Konstantinos Markopoulos$^{\star}$, Panos Kakoulidis$^{\star}$, Alexandra Vioni$^{\star}$, Myrsini Christidou$^{\star}$, } \\
	{\bf Junkwang Oh$^{\dagger}$, Gunu Jho$^{\dagger}$, Inchul Hwang$^{\dagger}$, Georgios Vardaxoglou$^{\star}$, }\\
	{\bf Aimilios Chalamandaris$^{\star}$, Pirros Tsiakoulis$^{\star}$, Spyros Raptis$^{\star}$} \\
	$^{\star}$ Innoetics, Samsung Electronics, Greece \\
	$^{\dagger}$ Mobile eXperience Business, Samsung Electronics, Republic of Korea}

\begin{document}
\maketitle
\begin{abstract}
Emotion detection in textual data has received growing interest in recent years, as it is pivotal for developing empathetic human-computer interaction systems.
This paper introduces a method for categorizing emotions from text, which acknowledges and differentiates between the diversified similarities and distinctions of various emotions. 
Initially, we establish a baseline by training a transformer-based model for standard emotion classification, achieving state-of-the-art performance. 
We argue that not all misclassifications are of the same importance, as there are perceptual similarities among emotional classes.
We thus redefine the emotion labeling problem by shifting it from a traditional classification model to an ordinal classification one, where discrete emotions are arranged in a sequential order according to their valence levels.
Finally, we propose a method that performs ordinal classification in the two-dimensional emotion space, considering both valence and arousal scales. 
The results show that our approach not only preserves high accuracy in emotion prediction but also significantly reduces the magnitude of errors in cases of misclassification.

\end{abstract}

\section{Introduction}

Emotion prediction from textual data has increasingly become important in natural language processing (NLP), as it 
lays the foundations for interactive and personalized computing; 
from enhancing the empathetic responses of chatbots to providing emotion-aware prompts in text-to-speech (TTS) systems. 
The ability to accurately infer emotional states from text remains challenging, due to the absence of relevant cues which are only present in speech, such as tone and pitch.
Emotions are not always explicitly stated in the text, and intended emotion may be classified ambiguously, even by humans. 
Traditional classification models treat emotions as discrete classes, offering a binary or multi-class output that may not fully capture the spectrum of human emotions \cite{demszky2020goemotions, kumar2022bert, abas2022bert, safaya2020kuisail, cortiz2021exploring, koufakou2022automatically}.
In this paradigm, the model does not account for the similarities among classes, e.g. the misclassification of sadness for joy is equivalently wrong as that of sadness for depression.
In downstream applications like TTS, such errors can lead to a substantial misrepresentation of the intended emotional tone and an unnatural outcome, e.g. uttering sad content with an excited voice.

\subsection{Related Work}

In recent years, transformer-based models have emerged as state-of-the-art in text analysis research. 
Models such as BERT \cite{devlin2018bert}, RoBERTA \cite{liu2019roberta} and XLNet \cite{yang2019xlnet} are pre-trained on large corpora in an unsupervised manner, and leverage contextual representations to model the natural language. 

BERT has been used for the tasks of sentiment analysis and emotion recognition of Twitter data with the addition of classifiers \cite{chiorrini2021emotion}.
For multi-class textual emotion detection, a CNN layer has been utilized to extract textual features and a BiLSTM layer to order text and sequence information \cite{kumar2022bert}.
Additionally, BERT has been leveraged to train a word-level semantic representation language model 
\cite{abas2022bert, safaya2020kuisail}. The semantic vector is then placed into the CNN to predict the emotion label. Results showed that BERT-CNN model overcomes the state-of-art performance.

The application of transformer-based models in emotion recognition has been investigated utilizing the GoEmotions dataset \cite{demszky2020goemotions}. RoBERTa demonstrated superior performance in comparison to the rest models \cite{cortiz2021exploring}.
Another study explored the performance of these models for emotion recognition on 3 datasets (GoEmotions, Wassa-21, and COVID-19 Survey Data) and confirmed the supremacy of RoBERTa~\cite{koufakou2022automatically}.
A Label-aware Contrastive Loss (LCL), which helps the model to differentiate the weights between different negative samples, has been recently introduced \cite{suresh2021not}. This enables the model to learn which pairs of classes are more similar and which differ.

In terms of representing emotions, the discrete emotional states may be mapped into ordinal scales in the two dimensions of valence and arousal, based on Russell’s circumplex model of affect \cite{russell1980circumplex}, as it has been applied already in real-valued data \cite{paltoglou2012seeing}.
In \cite{park2019dimensional} a model is used to predict emotions across valence, arousal, and dominance (VAD) dimensions, using a categorical emotion-annotated corpus and Earth Mover's Distance (EMD) loss. 
It achieves state-of-the-art performance in emotion classification and correlates well with ground truth VAD scores. 
The model improves with VAD label supervision and can identify emotion words beyond the initial dataset.

\subsection{Contribution}
In this work, we introduce an emotion-classification method that achieves state-of-the-art performance while accounting for the perceptual distance of emotional classes according to Russell's circumplex model of affect.
First, 
we establish a RoBERTa-CNN baseline model, which achieves similar performance to existing transformer-based models on standard emotion classification tasks.
That model is then adapted for ordinal classification, where discrete emotions are arranged in a sequential order according to their valence. 
Finally, 
we propose ordinal classification in the two-dimensional emotion space, considering both valence and arousal scales. 
We prove that this approach not only maintains high classification accuracy, but also provides more meaningful predictions in cases of misclassifications.

This paper does not aim to introduce a novel model architecture for the task of emotional classification. 
We adopt established model architectures, that have already demonstrated high efficiency, and focus on minimizing the effect of errors in emotion classification. 
Therefore, the contributions of this study are outlined as follows:
\begin{itemize}
	\item Propose an ordinal classification method for emotion prediction from text that achieves the same accuracy and F1 score of other state-of-the-art approaches.
	\item Show that with this method the model makes less severe mistakes.
	\item Enhance the capabilities of the model to perform emotion classification for a wide variety of emotions by introducing ordinal classification in the 2D space using the valence and arousal scales.
\end{itemize}

\section{Data}
We used the ISEAR, Wassa-21 and GoEmotions datasets in our study,
which are publicly available and are commonly used in relevant works.

ISEAR 
dataset~\cite{scherer1990international} is a balanced dataset constructed through cross-culture questionnaire studies. 
It contains 7666 sentences classified into seven distinct emotion labels: joy, anger, sadness, shame, guilt, surprise, and fear.

Wassa-21 
was part of the WASSA5 2021 Shared Task on Empathy Detection and Emotion Classification. 
The dataset contains essays in which authors expressed their empathy and distress in reactions to these news articles. 

GoEmotions 
was presented in ~\cite{demszky2020goemotions}. 
The original dataset contains about 58k Reddit comments with human annotations mapped into 27 emotions or neutral. 

To make our model comparable to other approaches we pre-processed our datasets following ~\cite{adoma2020comparative} for ISEAR and ~\cite{koufakou2022automatically} for Wassa-21 and GoEmotions keeping only that follows Ekman’s emotions \cite{ekman1992argument}.

\begin{figure*}[tp]
	\setlength{\belowcaptionskip}{-7pt}
	\centering
	\includegraphics[width=0.7\linewidth]{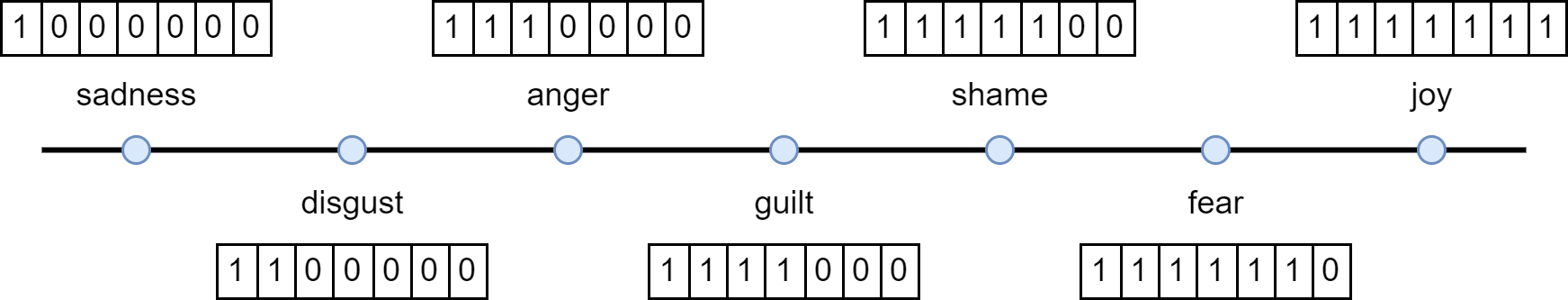}
	\caption{ISEAR Emotions Valence Order}
	\label{fig:ordinalemotions}
\end{figure*}

\section{Baseline Model}
Initially, our objective was to develop a baseline model that could perform competitively with state-of-the-art benchmarks.
We developed a RoBERTa-CNN model for emotion classification as it provides better results than the standard baselines Table \ref{tab:metrics}. 
Text classification models commonly adopt a two-part structure, consisting of: $1)$ the transformer-based model and $2)$ the classification head.
Prior research has extensively compared foundational transformer-based models in the context of text classification tasks, proving that RoBERTa outperforms others as an enhanced iteration of BERT with a larger pre-trained corpus.
Our initial experimentation involving BERT, RoBERTa, DistilBERT, and XLnet, verified this conclusion.

\begin{table}[]
	\setlength{\tabcolsep}{2pt}
	\setlength{\extrarowheight}{7pt}
	\begin{subtable}[]{\linewidth}
		\centering
		\begin{tabular}{lccc}
			\textbf{}          & \textbf{GoEmotions} & \textbf{Wassa-21} & \textbf{ISEAR} \\ \hline\hline
			\makecell[l]{Previous \\ [-3pt] best\footnote{\cite{koufakou2022automatically}, \cite{adoma2020comparative}}}      & 0.83               & 0.54              & \textbf{0.74}           \\
			\makecell[l]{Proposed\\[-3pt] baseline} & \textbf{0.85}               & \textbf{0.62}             & 0.73         \\
			\makecell[l]{Proposed\\[-3pt] ordinal} & \textbf{0.85}               & 0.56             & 0.73         \\ \hline
		\end{tabular}
		\label{tab:f1_scores}
		\caption[]{F1-score}
	\end{subtable}
	\begin{subtable}[]{\linewidth}
		\centering
		\begin{tabular}{lccc}
			\textbf{}          & \textbf{GoEmotions} & \textbf{Wassa-21} & \textbf{ISEAR} \\ \hline\hline
			\makecell[l]{Proposed\\[-3pt] baseline} & \textbf{0.85}               & \textbf{0.69} & \textbf{0.73} \\
			\makecell[l]{Proposed\\[-3pt] ordinal} & \textbf{0.85} & 0.68             & \textbf{0.73} \\ \hline
		\end{tabular}
		\label{tab:accuracy}
		\caption[]{Accuracy}
	\end{subtable}
	\vspace{-5pt}
	\caption[]{Evaluation metrics}
	\label{tab:metrics}
	\vspace{-5pt}
	\label{tab:metrics}
\end{table}

In constructing the baseline model, we conducted additional experiments focusing on the classification head.
Our classification head consists of two convolutional neural network (CNN) layers with kernel sizes [6,4] and [1024, 2048] the number of filters respectively. 
The encoded information is compressed using mean pooling and the resulting vector undergoes a 3-layer feedforward neural network (FFNN) [2048, 768, \#number\_of\_classes] with softmax in the end. 
\textit{Experiments followed these hyperparameters:  epochs=10, learning rate=0.6e-5, batch\_size=16, max\_seq\_length=200, AdamW optimizer.} 

Acknowledging that even with the state-of-the-art approaches, models inevitably commit errors, we have introduced an ordinal classification approach aimed at reducing significant misclassifications on emotion recognition task.

\section{Ordinal Classification}

Following the previous approach, we fine-tuned our model utilizing a standard cross-entropy loss where each label is discrete. 
An inherent limitation of the cross-entropy loss lies in its treatment of misclassifications as nominal rather than ordinal. In this context, misclassifying a “positive” as a “very positive” is no worse (in terms of loss) as “very negative”.
However, following this methodology is not optimal when we refer to emotions, e.g. misclassifying joy as excitement, is different from a misclassification to sadness.
To address this, we arrange the emotions in an ordinal manner based on their valence level as illustrated in Figure  \ref{fig:ordinalemotions}.

In order to minimize the gaps between labels in our model, we replaced the discrete one-hot representations of emotions with ordinal ones. By employing Mean Square Error (MSE) loss during training, our model focuses on narrowing the gap between target and prediction distances, emphasizing not only the correct classification but also the overall reduction of discrepancies.
We experimented further by using regression loss instead of ordinal loss, however, the initial results favored the latter.

Following the ordinal classification, our baseline model achieved competitive performance on the three datasets, having a quicker convergence in every case Table \ref{tab:metrics}. 
The main contribution here is that even if the overall performance does not change the misclassification error decreases.
By following this approach, there are fewer misclassifications between emotions that are distant and more between emotions that have similar valence.

On Wassa-21 dataset the ordinal model did not achieve a macro-F1 score comparable to the baseline, despite maintaining an equally high accuracy. 
This can be attributed to the fact that the dataset was unbalanced and MSE did not have a mechanism to handle it.
We further examined the ISEAR dataset for its balance, featuring a substantial number of examples for each emotion category. 

\begin{figure}

		\subfloat[Baseline\label{fig:confusionmatrixsoft}]{\includegraphics[width=0.9\columnwidth]{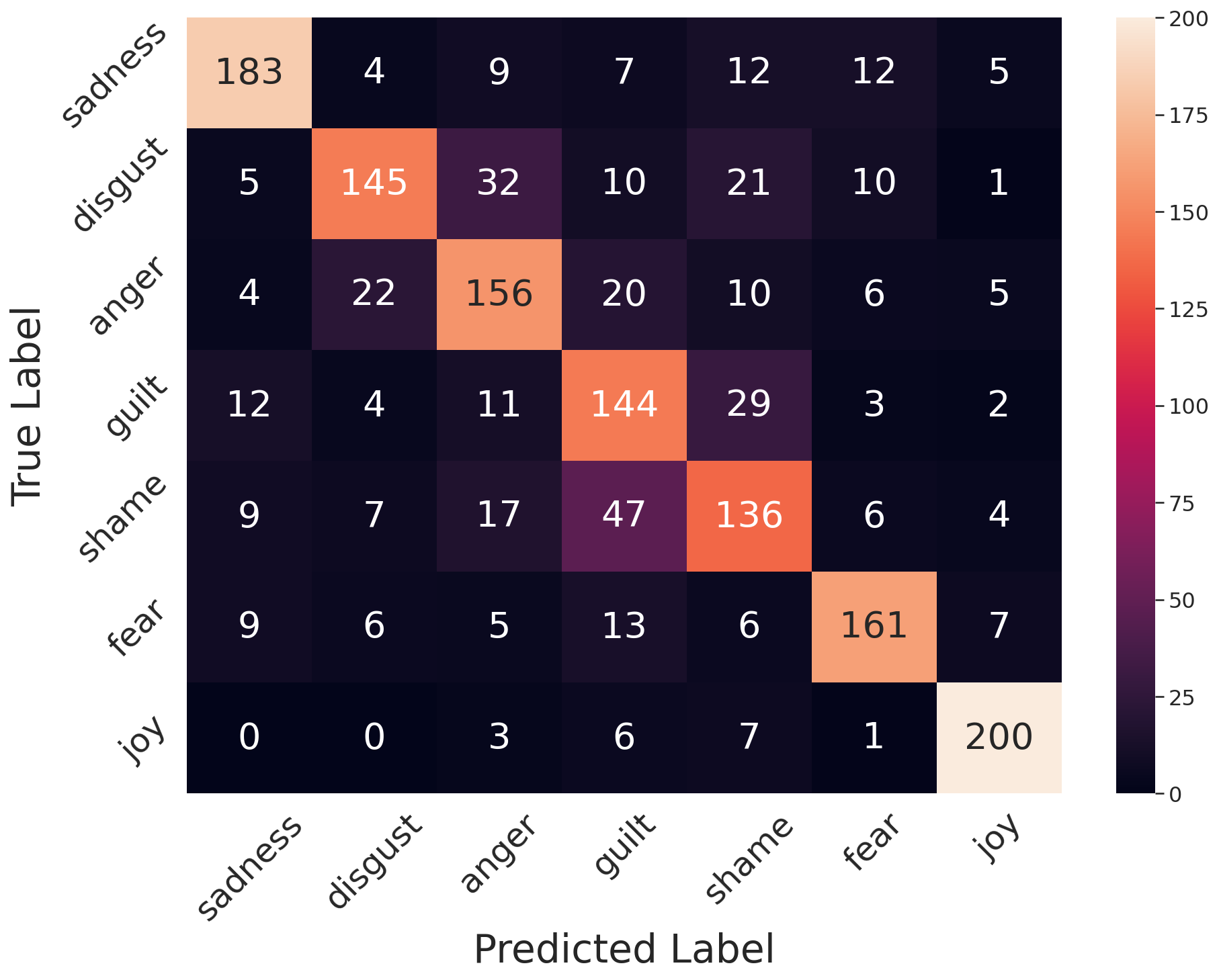}}
	
		\subfloat[Ordinal\label{fig:confusionmatrixordinal}]{\includegraphics[width=0.9\columnwidth]{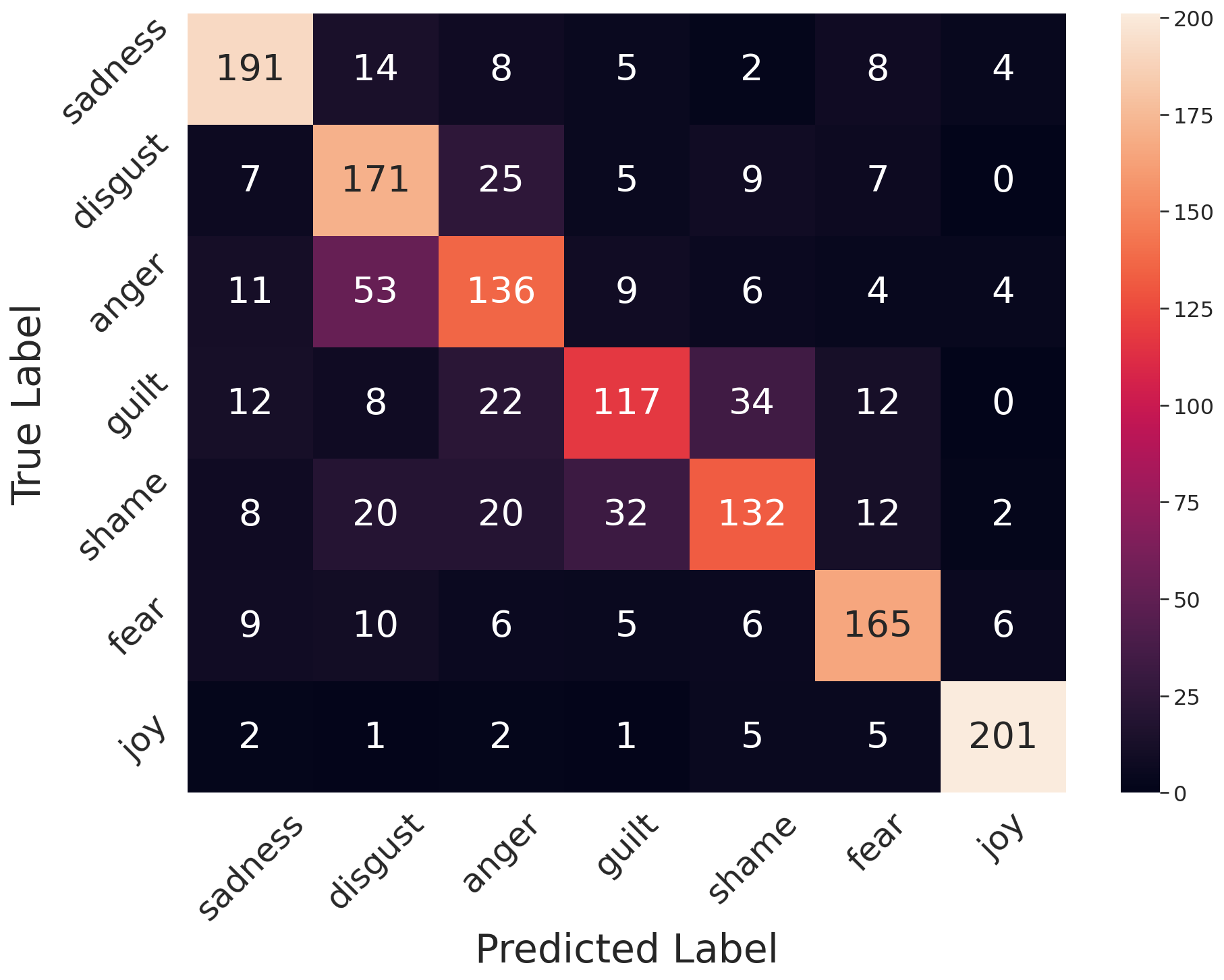}}

\caption{Confusion Matrices on ISEAR dataset}
\label{fig:confusionmatrix}
\end{figure}

\begin{figure}
	\setlength{\belowcaptionskip}{-7pt}
	\centering
	\includegraphics[width=0.8\linewidth]{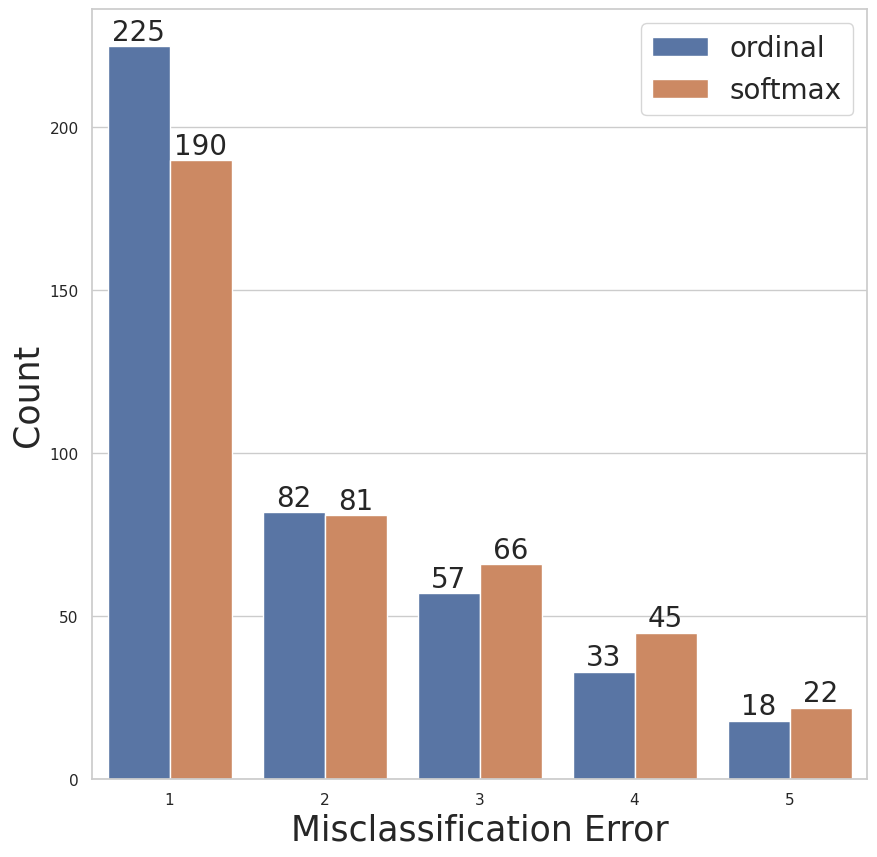}
	\caption{Error histograms of models trained for ordinal and baseline (softmax) classification on ISEAR dataset.}
	\label{fig:histordinal}
\end{figure}

The ordinal classification forces the model to make less severe mistakes, by penalizing higher misclassifications that are very far from the ground truth regarding the valence order.
Even if the accuracy and F1-score are similar to the base model, the effectiveness of ordinal can be seen through the confusion matrices in Figures \ref{fig:confusionmatrix} and error distances histogram in Figure~\ref{fig:histordinal}. In the first case, the baseline confusion matrix (Figure \ref{fig:confusionmatrixsoft}) makes more severe misclassifications that are far from the main diagonal. 
In contrast, on ordinal confusion matrix (Figure \ref{fig:confusionmatrixordinal}) the misclassifications tend to distant the upper right and the down left corners, where the misclassification error is max, and gather around the diagonal.
Moreover, this phenomenon can be observed through error distances histogram Figure~\ref{fig:histordinal}, in which we count the number of misclassification errors for each case. The misclassification error is defined as the distance between the target and the prediction on valence scale (i.e if the target was sadness and the prediction was anger the misclassification-error is 2 and if the prediction was fear the misclassification-error would be 5). The histograms show that the ordinal approach prefers to make misclassifications with distances of 1 rather than errors with distances larger than 3.

\section{2D Ordinal Classification}

\begin{figure}[h]
	\setlength{\belowcaptionskip}{-7pt}
	\centering
	\includegraphics[width=0.8\linewidth]{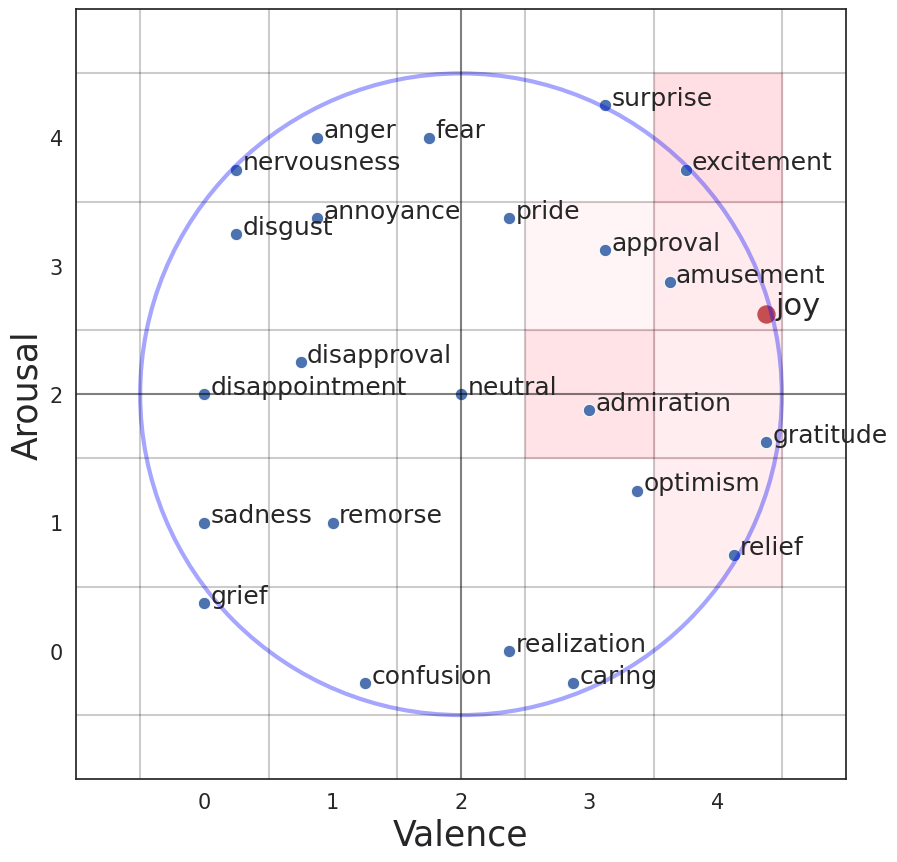}
	\caption{The emotions grid, as described by Russel. In pink color depicted the distribution of joy emotion, which was not seen during training.}
	\label{fig:joytaxonomy}
\end{figure}

However, expressing a broader range of emotions proves challenging when relying solely on valence levels, as certain emotions may share similar valence values (e.g., both excitement and amusement emotions describe a very positive state). 
To enhance the expressiveness of our model and encompass a wider variety of emotions we introduced a second dimension to our problem: the arousal scale.
Based on Russell's circumplex space model \cite{russell1980circumplex}, \cite{feldman1998independence}, we mapped a subset of 23 emotions to a 2D Cartesian coordinate system, where the emotions are represented as points and the x- and y-axis are valence and arousal, respectively \cite{scherer2005emotions} Figure~\ref{fig:joytaxonomy}.
To extend the ordinal approach on both dimensions, we separated the emotion space into a $5 \times 5$ grid space, where each emotion belongs in a unique cell (e.g., in Figure \ref{fig:joytaxonomy} grief and pride are mapped to $(0,0)$ and $(3,2)$ cells respectively).
We adapt our model for 2D classification task by maintaining the valence classifier and introducing a supplementary classifier head for predicting the arousal level of each emotion. 
Our model is trained to classify the given text in two manners, valence and arousal following the ordinal approach presented before. 
Both heads are trained simultaneously by combining their losses.
The anticipated valence and arousal levels serve as the coordinates within the emotion grid.

To evaluate our 2D ordinal approach we utilized GoEmotion dataset, which offers a broad spectrum of emotions. 
Among the 27 emotions available, we incorporated 23, ensuring that each grid cell corresponds to, at most, one emotion label.
Both approaches followed the previously outlined hyper-parameter set during training.
The results are presented in Table \ref{tab:2dordinal}. 
It is apparent that our baseline model struggled to effectively categorize all 23 distinct emotion labels.
Conversely, our 2D model combined with ordinal classification performs significantly better on this challenging task. 
In addition, employing ordinal classification enabled the model to discern similarities between emotions by minimizing the distances between target and prediction on both valence and arousal dimensions.
This is evident in Figure~\ref{fig:joytaxonomy}, where the model, even when lacking exposure to instances of the \textit{joy} emotion during training, accurately classifies input examples of joy in close proximity to the actual ground truth location for joy (depicted by the red dot) avoiding distant misclassifications.

\begin{table}[h]
	\centering
	\renewcommand*{\arraystretch}{1.1}
	\begin{tabular}{lcc}
		\textbf{}          & \multicolumn{2}{c}{\textbf{GoEmotions}} \\ \hline\hline
		& F1-score & Accuracy \\ \hline
		Proposed baseline & 0.12 & 0.28\\
		Proposed 2D ordinal & \textbf{0.63} & \textbf{0.52} \\ \hline
	\end{tabular}
	\caption[]{Classification metrics on 23 emotions}
	\label{tab:2dordinal}
\end{table}

\section{Conclusion}
In this paper we presented a novel approach to emotion prediction from textual data, recognizing the nuanced similarities and distinctions among various emotions. Initially, we introduced a RoBERTa-CNN model for standard emotion classification as our baseline. 
By arranging emotions based on valence levels we shifted from traditional classification to ordinal.
Further innovation introduces ordinal classification in the two-dimensional emotional space, considering both valence and arousal scales. 
The proposed methodology enhances the model's performance by providing more meaningful predictions, taking into account the correlations between emotions.

Future directions involve extending research to diverse datasets, exploring alternative models, and experimenting with different emotion ordering schemes.
An interesting direction also involves interpreting the model's components in order to better understand the importance of each feature in order to improve the existing method.

\bibliography{anthology,custom}

\appendix

\end{document}